\documentclass[runningheads]{llncs}

\usepackage{eccv}

\newcommand{\Tref}[1]{Table~\ref{#1}}
\newcommand{\Eref}[1]{Equation~(\ref{#1})}
\newcommand{\Fref}[1]{Fig.~\ref{#1}}
\newcommand{\Sref}[1]{Section~\ref{#1}}

\newcommand{\eg}[1]{\emph{e.g.}}

\usepackage{eccvabbrv}

\usepackage{graphicx}
\usepackage{booktabs}
\usepackage{wrapfig}
\usepackage{adjustbox}
\usepackage{multirow}
\usepackage{floatrow}

\usepackage[accsupp]{axessibility}  

\usepackage{hyperref}

\usepackage{orcidlink}

\begin{document}

\title{GeometrySticker: Enabling Ownership Claim of Recolorized Neural Radiance Fields} 

\titlerunning{Enabling Ownership Claim of Recolorized Neural Radiance Fields}

\author{Xiufeng Huang\inst{1,2} \and 
Ka Chun Cheung\inst{2} \and 
Simon See\inst{2} \and 
Renjie Wan\inst{1}\thanks{Corresponding author.}
}

\authorrunning{X.~Huang et al.}

\institute{Department of Computer Science, Hong Kong Baptist University \and
NVIDIA AI Technology Center, NVIDIA \\
\email{xiufenghuang@life.hkbu.edu.hk, \{chcheung, ssee\}@nvidia.com, renjiewan@hkbu.edu.hk}}

\maketitle

\begin{abstract}
Remarkable advancements in the recolorization of Neural Radiance Fields (NeRF) have simplified the process of modifying NeRF's color attributes. Yet, with the potential of NeRF to serve as shareable digital assets, there's a concern that malicious users might alter the color of NeRF models and falsely claim the recolorized version as their own. To safeguard against such breaches of ownership, enabling original NeRF creators to establish rights over recolorized NeRF is crucial. While approaches like CopyRNeRF have been introduced to embed binary messages into NeRF models as digital signatures for copyright protection, the process of recolorization can remove these binary messages. In our paper, we present GeometrySticker, a method for seamlessly integrating binary messages into the geometry components of radiance fields, akin to applying a sticker. GeometrySticker can embed binary messages into NeRF models while preserving the effectiveness of these messages against recolorization. Our comprehensive studies demonstrate that GeometrySticker is adaptable to prevalent NeRF architectures and maintains a commendable level of robustness against various distortions. 
Project page: \href{https://kevinhuangxf.github.io/GeometrySticker/}{https://kevinhuangxf.github.io/GeometrySticker}.

\end{abstract}

\section{Introduction}
\label{sec:intro}

Significant progress~\cite{Wang_Chai_He_Chen_Liao_2022, Kuang_Luan_Bi_Shu_Wetzstein_Sunkavalli_2022, gong2023recolornerf} has been made in the recolorization of Neural Radiance Fields~\cite{mildenhall2021nerf}, allowing people to adjust NeRF's color properties easily. However, as NeRF, a key representation of 3D scenes, becomes possible to be shareable digital assets~\cite{nerfstudio}, there is a risk that ill-intentioned users could recolorize NeRF models and illegitimately assert ownership over the modified versions. To prevent NeRF models from such ownership breaches,  it is important to allow original NeRF creators to claim ownership for a recolorized NeRF.

A convenient way to assert ownership of digital assets is to embed invisible binary messages as digital watermarking into digital assets. Then, once the digital assets are maliciously edited, the owners can extract binary messages from NeRF \cite{mildenhall2021nerf} to claim ownership. For example, CopyRNeRF~\cite{luo2023copyrnerf} has been introduced to safeguard the copyright of NeRF. Their method embeds binary messages in a way that aligns with the color representation within NeRF. Then, they utilize a message decoder to extract binary messages from images rendered from NeRF to verify the copyright. Yet, as CopyRNeRF~\cite{luo2023copyrnerf} deeply relies on the combination between binary messages and color representation, such messages directly become ``irretrievable'' on rendered images when simply altering the color representation. Thus, the cornerstone of ensuring ownership claim over the recolorized NeRF model lies in making binary messages robust to alternations to color representation.

If recolorization is mainly conducted on color representation, a straightforward solution would be to hide messages into cover media \cite{kalita2018steganography} unrelated to color representation. 
Then, the envisioned ownership claim and recolorization can be achieved in a concurrent way. In general, NeRF \cite{mildenhall2021nerf} relies on color and geometry to finish volume rendering. As the two key components are represented separately via two MLPs, geometry components become suitable cover media for hiding binary messages. Then, we can balance the target for claimability and recolorization.


\begin{figure}[t]
  \centering
  \includegraphics[width=\linewidth]{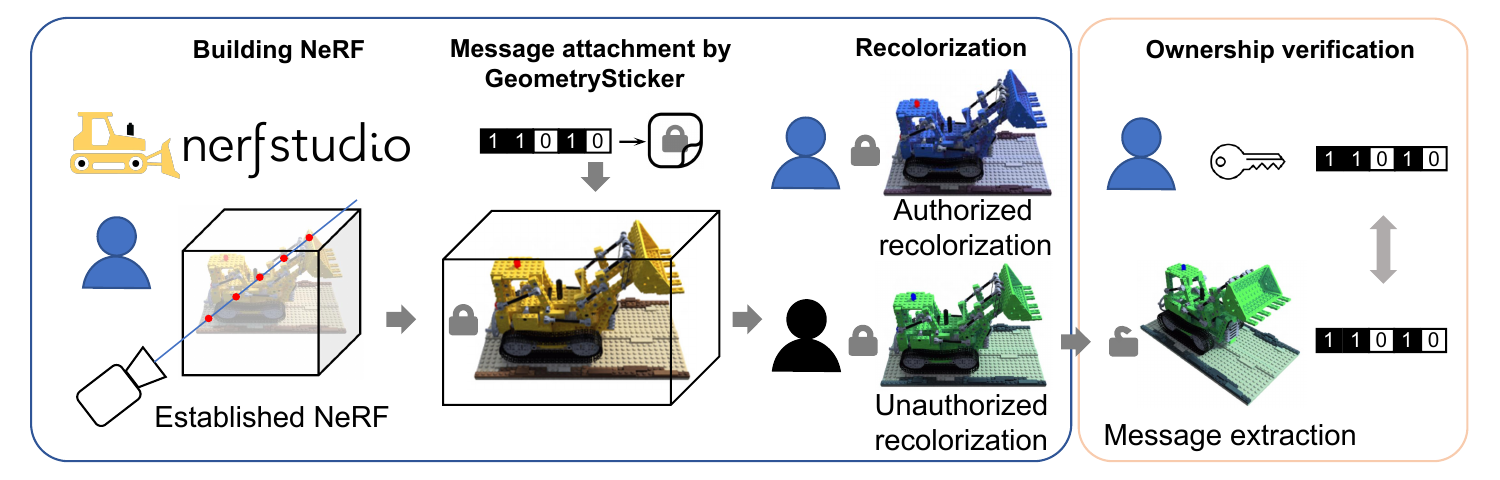}
   \caption{Our proposed scenario for ownership claim over the recolorized NeRF. Users can construct their NeRF models using readily available platforms, such as NeRFStudio. \textbf{Ownership message attachment}: They can then swiftly stick binary messages onto those created NeRF models via our proposed GeometrySticker. These watermarked NeRF models remain suitable for standard recolorization processes (herein termed Authorized recolorization). \textbf{Ownership verification}: Should unauthorized recolorization occur, the creators of the NeRF models can retrieve the watermarks from the altered models to verify ownership.
   }
   
   \label{fig:proposed_scene}
\end{figure}

However, three challenges stand in the way of achieving the above goals. \textbf{First}, since its introduction, several NeRF variants~\cite{yu_and_fridovichkeil2021plenoxels, mueller2022instant, Chen2022ECCV} have already been developed. Their internal structures used for geometry representation also have significant differences. 
\textit{The watermarks should be scalable to NeRF variants.}
\textbf{Second}, 
NeRF \cite{mildenhall2021nerf} and its variations \cite{ mueller2022instant, Chen2022ECCV} utilize neural networks to learn implicit neural representations for reconstructing 3D scenes.
Additional messages can easily disrupt the geometry structure represented in the networks. 
Then, it is essential to ensure that \textit{the changes made by the embedded binary messages are minimal}. 
\textbf{At last}, 
an effective watermarking method should extract the copyright messages from such minimal changes.

We introduce GeometrySticker for the ownership claim of recolorized NeRF. GeometrySticker is with the following characteristics to address the above challenges: 1) \textbf{Scalability}. Rather than embedding messages into geometry representation, we propose to attach messages onto geometry like stickers. Besides a simple network used for message representation, the attachment of such a sticker does not rely on any additional structure-specific optimization, which can be scalable to various variants. 
2) \textbf{Subtlety}. The cover media selected for embedding binary messages occupy only small sections within the NeRF model, ensuring that modifications introduced by the message embedding remain subtle. 3) \textbf{Ubiquity}. The cover media are accessible from every viewpoint, ensuring NeRF owners can retrieve binary messages from each perspective.

\begin{figure*}[t!]
  \centering
  \begin{subfigure}{\linewidth}
    \includegraphics[width=\linewidth]{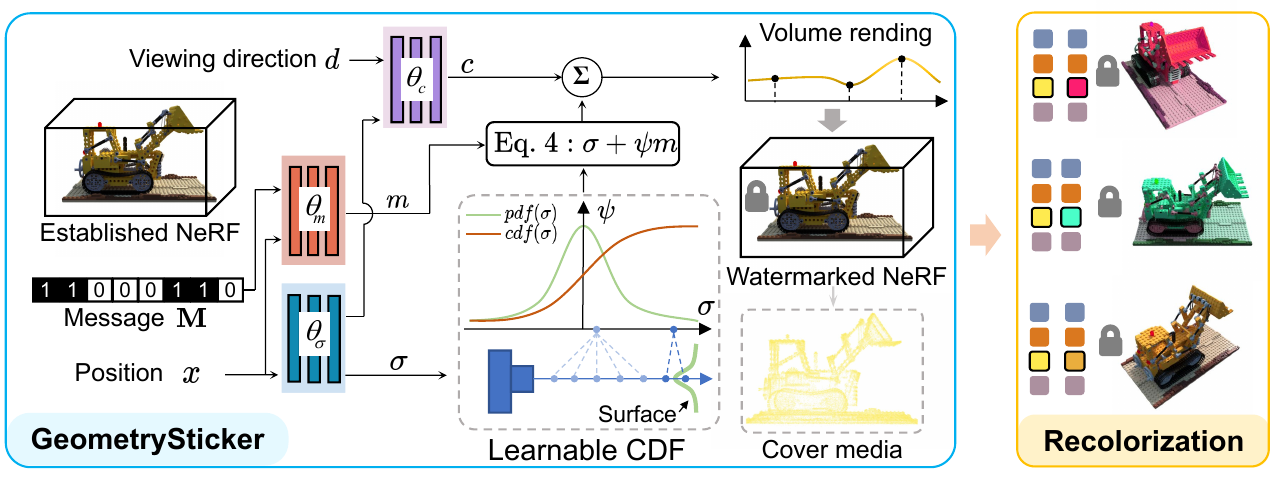}
  \end{subfigure}
  
  \caption{The framework of our proposed GeometrySticker. We employ a Multilayer Perceptron (MLP), denoted as $\theta_{m}$ to convert binary messages into a format that aligns with NeRF's geometry representation. Following this, a learnable Cumulative Distribution Function (CDF) is utilized to select 3D points close to the object surfaces with high geometry values as the cover medium. Subsequently, an addition is applied to attach the message (\Eref{eq:encode_watermark}) onto the chosen cover media. Through this message attachment process, a watermarked NeRF is generated, capable of retaining its efficacy across diverse recolorizations. Should any unauthorized changes to color attributes occur, NeRF owners can retrieve the integrated watermarks to assert their ownership. $\theta_{\sigma}$ and $\theta_c$ indicate the representation for geometry and color.}
  
  
  \label{fig:algo_overview}
\end{figure*}

In \Fref{fig:algo_overview}, we display our framework for implementing GeometrySticker. Our GeometrySticker is a lightweight Multilayer Perceptron (MLP) capable of translating binary messages into formats compatible with the geometry representation used in NeRF and its variants~\cite{yu_and_fridovichkeil2021plenoxels, mueller2022instant, Chen2022ECCV, barron2021mip}. Then, we can seamlessly integrate this compatible form of message representation with the chosen cover media, irrespective of the NeRF architectures, thus guaranteeing scalability. Then, we employ the 3D points sampled throughout the ray marching process as the cover media, and we specifically choose the 3D points near the objects' surfaces to affix our binary messages. These selected 3D points around the objects' surfaces represent a minor fraction of the total representations, thus ensuring subtlety. Ultimately, we guarantee that these cover media are accessible from every viewpoint, ensuring that the messages can be accessed from each perspective, thereby ensuring ubiquity.
  
As shown in \Fref{fig:proposed_scene}, after the message attachment, we can easily recolorize the watermarked NeRF with GeometrySticker for authorized recolorization. However, if unauthorized recolorization is triggered, NeRF creators can easily retrieve binary messages from 2D images rendered from recolorized NeRF for ownership verification. The proposed GeometrySticker has the following key characteristics:

\begin{itemize}


    \item Safeguarding the ownership claim of NeRF even when the color attributes have been altered
    
    \item Using the geometry components associated with selected 3D points to achieve the subtlety and ubiquity of the embedded ownership messages. 
    
    
    \item Designing a message sticker for ownership message attachment to achieve the scalability of our proposed solution.

\end{itemize}

Our GeometrySticker, exhibiting high scalability, is capable of being generalized across various NeRF variants~\cite{yu_and_fridovichkeil2021plenoxels, mueller2022instant, Chen2022ECCV, barron2021mip} that use neural representations for geometry. Based on our experiments, the use of GeometrySticker does not impair the effectiveness of current recolorization approaches \cite{Wang_Chai_He_Chen_Liao_2022, Kuang_Luan_Bi_Shu_Wetzstein_Sunkavalli_2022, gong2023recolornerf} designed for NeRF. Furthermore, besides the recolorization approaches for NeRF, the binary messages can still be reliably extracted even when the 2D images rendered from NeRF are subjected to direct image-level color modifications.

\section{Related work}
\label{sec:related_work}

\noindent\textbf{NeRF and its variants.} NeRF \cite{mildenhall2021nerf} and its variants \cite{yu_and_fridovichkeil2021plenoxels, mueller2022instant, barron2021mip, zhu2022neural, zhu2023occlusion, tang2024neural} show the capability to create realistic 3D representations of objects and scenes from 2D images with different perspectives. To improve the efficiency of scene representation, Plenoxel \cite{yu_and_fridovichkeil2021plenoxels} reconstructs the scene in a sparse voxel grid and renders each ray sample via trilinear interpolation of the neighboring voxel coefficients. TensoRF \cite{Chen2022ECCV} factorizes the radiance fields by vector-matrix decomposition for efficient scene modeling. InstantNGP \cite{mueller2022instant} optimizes the input encoding with a multi-resolution hash table to reduce the number of floating point and memory access operations. 
As the creation of NeRF becomes more accessible, individuals are more inclined to easily recreate their preferred 3D scenes and share them with the public. It's important to address potential breaches of ownership in the process of such sharing.

\noindent\textbf{Recolorization of NeRF.} The recolorization of NeRF \cite{Kuang_Luan_Bi_Shu_Wetzstein_Sunkavalli_2022, Wang_Chai_He_Chen_Liao_2022, gong2023recolornerf, cheng2024colorizing} has achieved remarkable performance. 
CLIP-NeRF \cite{Wang_Chai_He_Chen_Liao_2022} use text prompts to alter the color supervised by CLIP-based \cite{radford2021learning} matching loss. PaletteNeRF\cite{Kuang_Luan_Bi_Shu_Wetzstein_Sunkavalli_2022} decomposes the appearance of 3D points into a linear combination of palette-based bases across the scene for photorealistic color editing. 
Similarly, RecolorNeRF\cite{gong2023recolornerf} also decomposes the scene into a set of color layers to form a palette for color altering based on the TensoRF \cite{Chen2022ECCV} architecture.
As the recolorization of NeRF has become effective in recolorizing the 3D scene, it is important to explore an effective approach to protect the intellectual property of NeRF models when their color properties are modified. 

\noindent\textbf{Ownership assertion of NeRF.} 
Traditional 2D watermarking methods embed information in the least significant bits of image pixels \cite{van1994digital}. Significant progress has been made in deep-learning-based image watermarking \cite{tancik2020stegastamp, weng2019high, wengrowski2019light, yang2021robust, zhang2020udh, zhang2019robust, wang2024spy}. HiDDeN \cite{zhu2018hidden} is one of the first deep image watermarking methods that outperforms traditional methods. 
3D watermarking approaches are usually designed for explicit 3D models~\cite{ohbuchi2002frequency, praun1999robust, wu20153d, chen2023mimic3d, son2017perceptual, yoo2022deep}.
However, these methods are not applicable to the copyright protection of NeRF due to its implicit property. 
Recently, StegaNeRF \cite{li2023steganerf} designs a framework for steganographic information embedding in NeRF renderings. CopyRNeRF \cite{luo2023copyrnerf} generates watermarked color representations to ensure the invisibility of hidden copyright messages. However, when faced with recolorization, the hidden information embedded in the two methods might become unrecoverable. This prompts us to investigate ensuring ownership claims over recolorized NeRF models.


\section{Preliminaries on the watermarking of NeRF}
\label{sec:preliminaries}

The goal of watermarking NeRF is to safeguard its copyright and ownership by integrating binary messages into this burgeoning digital asset for 3D scenes. In general, NeRF \cite{mildenhall2021nerf} builds a function $f$: $(\mathbf{x}, \mathbf{d}) \rightarrow(\sigma, \mathrm{c})$ to map the position $\mathbf{x}$ and viewing direction $\mathbf{d}$ to the point's density $\sigma$ and color $\mathrm{c}$. 
Vanilla NeRF\cite{mildenhall2021nerf} uses a MLP $\Theta_\sigma$ and the encoding function $\gamma_\mathbf{x}$ to map the 3D location $\mathbf{x}$ into density value $\sigma$ and the intermediate geometry feature output of $\mathbf{z}$: $[\sigma, \mathbf{z}]=\Theta_\sigma\left(\gamma_{\mathbf{x}}(\mathbf{x})\right)$.
Another MLP $\Theta_{c}$ and the encoding function $\gamma_\mathbf{d}$ are used to map the geometry feature $\mathbf{z}$ and viewing direction $\mathbf{d}$ into the color value $\mathbf{c}=\Theta_c\left(\mathbf{z}, \gamma_{\mathbf{d}}(\mathbf{d})\right)$. Once the optimization settles down, an implicit scene representation can be obtained, and all scene information can be stored in MLP as its network weights. 

Given the challenge of retrieving binary messages from the implicit representation of NeRF, existing strategies~\cite{luo2023copyrnerf, li2023steganerf} typically design a method to convey copyright messages from the implicit neural representation to the 2D rendered images. Following that, a CNN-based message extractor is commonly employed to retrieve binary messages from the 2D images. The binary message embedding can be represented as $f_{\mathbf{M}}$, where $\mathbf{M}$ is the binary messages with length $N_{\mathbf{M}}$. The message embedding is fused with the implicit neural representation for rendering the 2D watermarked images $\mathbf{I}_{w}$.
The corresponding message extraction process can be denoted as $\hat{\mathbf{M}}=D(\mathbf{I}_{w})$, where $D$ denotes a CNN-based message extractor used to extract the hidden information $\hat{\mathbf{M}}$ from the 2D watermarked image $\mathbf{I}_{w}$.


\section{Proposed method}
\label{sec:proposed_method}

We have shown the scenario for our GeometrySticker in \Fref{fig:proposed_scene}, where people can claim ownership over the recolorized NeRF. In our scenario, for a NeRF model established via public resources such as NeRFStudio~\cite{nerfstudio}, creators can effortlessly integrate binary messages into these established NeRF models using GeometrySticker for potential ownership claims. Once ownership messages are embedded into NeRF models, authorized recolorization can still be easily performed on the watermarked NeRF for legitimate uses. Yet, in the event of unauthorized recolorization, NeRF creators can utilize their available message extraction tool to retrieve binary messages from the rendered images, thereby affirming their ownership of their digital assets.



\subsection{GeometrySticker}

\noindent\textbf{Choosing cover media.}  Choosing suitable cover media to conceal ownership messages is crucial in digital watermarking~\cite{kalita2018steganography}. CopyRNeRF \cite{luo2023copyrnerf} has shown that embedding binary messages directly into geometry components can readily result in visible artifacts, thereby degrading the quality of scene representation. In fact, these artifacts partly originate from 3D points in empty spaces that have lower geometry values. In volume rendering~\cite{levoy1990efficient}, $N_{p}$ points are sampled along the camera marching rays with color and geometry values $\left\{\left(\mathbf{c}_i, \sigma_i\right)\right\}_{i=1}^{N_{p}}$. 
These points exhibit low geometry values in empty spaces and high values at the surfaces of objects~\cite{kajiya1984ray}. 
Incorporating additional information into low geometry value 3D points located in empty spaces can lead to easily detectable alterations.
Thus, rather than directly incorporating the hidden messages into the whole geometry representation, we consider those 3D points located on the object surfaces as the cover media. These 3D points on the object surfaces often exhibit high values, making the attachment of messages into them result in less conspicuous changes. 

To precisely pinpoint the 3D points located on the surfaces of objects, we utilize the Laplace Cumulative Distribution Function (CDF) equipped with a learnable parameter. This approach can help in determining an appropriate threshold to filter out 3D points that exhibit high geometry values as:
\begin{equation}
\psi = \frac{1}{2} + \frac{1}{2} \cdot \text{sign}(\sigma - \mu) \cdot \left(1 - \exp\left(-\frac{|\sigma - \mu|}{\beta}\right)\right),
\label{eq:learnable_cdf}
\end{equation}
\noindent where $\mu$ and $\beta$ are the average and deviation of the geometry field, and $\psi \in [0,1]$ is a probability that any geometry values in the geometry field is less than or equal to the given geometry values $\sigma$. The probability $\psi$ can be used as an importance value to indicate whether the geometry value is a large number or not.

While utilizing fixed parameters for thresholding points on object surfaces is feasible, adopting a naive strategy for determining mean and standard deviation values for such thresholding might lead to inflexible cutoffs. This could unintentionally cover too many points, causing notable distortions, or on the flip side, too few points, thus hindering the efficient embedding of messages for future extraction by a message extractor.

Rather than using a fixed threshold for cover media generation, we consider $\beta$ to be a learnable parameter to be optimized during the cover media generation. Then, the range used for the cover media generation can be adaptively adjusted according to rendered contents and message attachment efficiency to ensure the invisibility of embedded messages. The outcomes illustrated in \Fref{fig:algo_overview} demonstrate that the chosen 3D points, serving as the cover media, effectively form sparse point clouds that capture the essential information of the target objects. We optimize the importance value $\psi$ with a sparsity loss \cite{Lombardi_Simon_Saragih_Schwartz_Lehrmann_Sheikh_2019} as follows:
\begin{equation}
\mathcal{L}_{sparse}=\frac{1}{|N_{p}|} \sum_{\psi_i}\left[\log \left(\psi_i\right)+\log \left(1-\psi_i\right)\right],
\end{equation}
which forces the importance value $\psi$ to be close to either zero or one. The importance values $\psi$ close to one indicate the 3D points with high geometry values. These points are selected as the cover media and only occupy small sections of the NeRF geometry to ensure subtlety.

\noindent\textbf{Message sticker.} To maintain scalability, we avoid the data hiding techniques used in previous methods, such as CopyRNeRF \cite{luo2023copyrnerf}. 
CopyRNeRF \cite{luo2023copyrnerf} inherently alters the underlying NeRF structure, which depends on particular configurations and exhibits reduced scalability.
Instead, we propose a message sticker that can attach messages to the selected cover media by \textbf{summation} like a sticker.
The message sticker $\Theta_{\mathbf{m}}$ can be achieved via an MLP as:
\begin{equation}
    m =\Theta_{\mathbf{m}}(\gamma_{x}(\mathbf{x}), \mathbf{M}),
    \label{eq:message_sticker}
\end{equation}
where $\mathbf{M}$ is the binary message with length $N_{b}$ and $m$ is the one dimensional message embedding. 
Then, we can directly attach the message embedding $m$ into geometry $\sigma$ via $\psi$ defined in \Eref{eq:learnable_cdf}:
\begin{equation}
\mathbf{\Tilde{\sigma}} = \mathbf{\sigma} + \mathbf{\psi}m,
\label{eq:encode_watermark}
\end{equation}
where $\Tilde{\sigma}$ is a watermarked geometry value. During volume rendering, the information attached via the message sticker can be incorporated into the rendered pixel values $\Tilde{C}$ as follows:
\begin{equation}
\Tilde{C}=\sum_{i=1}^N\exp(-\sum_{j=1}^{i-1}\Tilde{\sigma}_j\delta_j)(1-\exp(-\Tilde{\sigma}_i\delta_i))\mathbf{c}_i,
\label{eq:watermarked_color_predict}
\end{equation}
where $\delta$ is the distance between adjacent sample points, $\mathbf{\Tilde{\sigma}}$ is the watermarked geometry with a message attachment and $c$ is the color sampled along the ray. 
During training, to guarantee ubiquity, we repeat the above operations in each viewing point, which ensures that such 3D points exist at each perspective.
During message extraction, the binary messages incorporated on cover media can be easily extracted from the watermarked image $\mathbf{I}_{w}$ via a message extractor $D_\chi$ as $\hat{\mathbf{M}} = D_\chi(\mathbf{I}_{w})$, where $\hat{\mathbf{M}}$ is the binary messages extracted by the message extractor and $\chi$ is a trainable parameter.


\noindent\textbf{Optimization}. The message attachment in~\Eref{eq:encode_watermark} is based on simple addition operation, with the training just focused on enabling the message sticker to adapt binary messages into a form that aligns with the diverse architectures employed by NeRF \cite{mildenhall2021nerf} and its variants. 
Our optimization contains three components: 1) we optimize the learnable variable $\beta$ defined in \Eref{eq:learnable_cdf} to find an appropriate threshold for identifying 3D points with high geometry values as the cover media; 2) we train the message sticker in \Eref{eq:message_sticker} for embedding the binary messages into the 3D points; 3) we also train the message extractor $D_{\chi}$ to extract the hidden message from the watermarked rendered images $\mathbf{I}_w$.
The above three components can be simply optimized via a message loss as 
$\mathcal{L}_{msg}=\text{BCE}\left(D_\chi(\mathbf{I}_{w}), \mathbf{M}\right)$, 
where $\text{BCE}$ is the binary cross entropy loss, $\mathbf{I}_{w}$ is the watermarked rendered image and $\mathbf{M}$ is the ground truth message. Then, the classical MSE loss adopted by NeRF~\cite{mildenhall2021nerf} is employed to ensure that the GeometrySticker does not compromise the visual quality of rendered contents as $\mathcal{L}_{cont} = \|\mathbf{I}_{w} - \mathbf{I}_{o}\|_2^2$, where $\mathcal{L}_{cont}$ is the content loss and $\mathbf{I}_{o}$ is the original image. 
We also train a CNN-based classifier $C_{\phi}$ to classify whether the rendered images contain watermarks as $\mathcal{L}_{cls} = \text{BCE}(C_{\phi}(\mathbf{I}_{w}), C_{\phi}(\mathbf{I}_{u}))$, where $\mathbf{I}_{u}$ is the unwatermarked rendered image and $\phi$ is a trainable parameter.
The overall loss functions for our GeometrySticker can be incorporated by optimizing the following loss functions:
 \begin{equation}
     \mathcal{L}_{total}=\mathcal{L}_{cont} + \mathcal{L}_{msg} + \mathcal{L}_{{cls}} + \mathcal{L}_{{sparse}}.
     \label{eq:total_loss}
 \end{equation}

\subsection{Recolorization} 
\label{ssec:recolorization}

Once the binary messages have been attached to the geometry components via GeometrySticker, watermarked NeRF models can be easily recolorized in a claimable manner via off-the-shelf approaches for recolorization. We consider two off-the-shelf recolorization approaches. The first one is CLIP-based recolorization proposed in CLIPNeRF~\cite{Wang_Chai_He_Chen_Liao_2022}. In CLIP-based recolorization, the color representation within a NeRF model is modified using CLIP \cite{radford2021learning} features derived from a specified text prompt. We adhere to the protocols set forth in CLIPNeRF \cite{Wang_Chai_He_Chen_Liao_2022}, employing a CLIP \cite{radford2021learning} feature loss to guide the update of the color representation. Our second recolorization strategy is the palette-based recolorization. This approach begins by establishing a color palette that encompasses all fundamental color components. Subsequent precise recolorization is accomplished by adjusting the color palette, specifically by allotting distinct RGB values to designated color layers. 
We randomly select $10$ reference colors from the Standard sRGB / Rec.$709$ color gamut to recolorize the NeRF model.
Additionally, users have the option to apply image-level recolorization techniques to directly alter the colors in rendered images using traditional methods (\eg, color jittering). 
Some recolorization results are displayed in \Fref{fig:nerf_recoloring}.
Due to page limitations, \textbf{more details about this part can be found in the supplementary materials}.

\subsection{Implementation details}
\label{ssec:method_implementation_details}

Our GeometrySticker is implemented on PyTorch. The whole pipeline can be easily combined with popular NeRF architectures like InstantNGP \cite{mueller2022instant} or TensoRF \cite{Chen2022ECCV}. Besides, we also implement on vanilla NeRF \cite{mildenhall2021nerf}. NeRF and InstantNGP \cite{mueller2022instant} utilize MLP layers, and TensoRF \cite{Chen2022ECCV} utilizes a density grid to predict volume density. 
We train our GeometrySticker to find the important geometry component for attaching the copyright messages.

The patch size is set to $400 \times 400$ for each rendered image during training. 
During training, we apply several types of 2D distortions on the watermarked rendered images to achieve robustness, including Gaussian noise, random rotation, random cropout, and Gaussian blur. 
As our motivation is to protect the copyright of NeRF that has already been created, we first create several NeRF models by training them on Blender and LLFF datasets \cite{mildenhall2021nerf, mueller2022instant} following standard settings. Then, we use GeometrySticker to attach binary messages on established NeRF and apply the aforementioned recolorization methods in \Sref{ssec:recolorization} on the NeRF watermarked by GeometrySticker to test the watermarking robustness against recolorizations. 
Specifically, we employ the VGG16 \cite{simonyan2014very} network as the backbone of the CNN-based message extractor. An average pooling is then performed, followed by a final linear layer with a fixed output dimension $N_{b}$ to produce the continuous predicted message $\hat{\mathbf{M}}$. Training typically takes 5000 steps and can be completed within 45 minutes. All experiments are conducted on a single V100 GPU. 

\begin{figure}[t!]
  \centering
  \begin{subfigure}{\linewidth}
    \includegraphics[width=\linewidth]{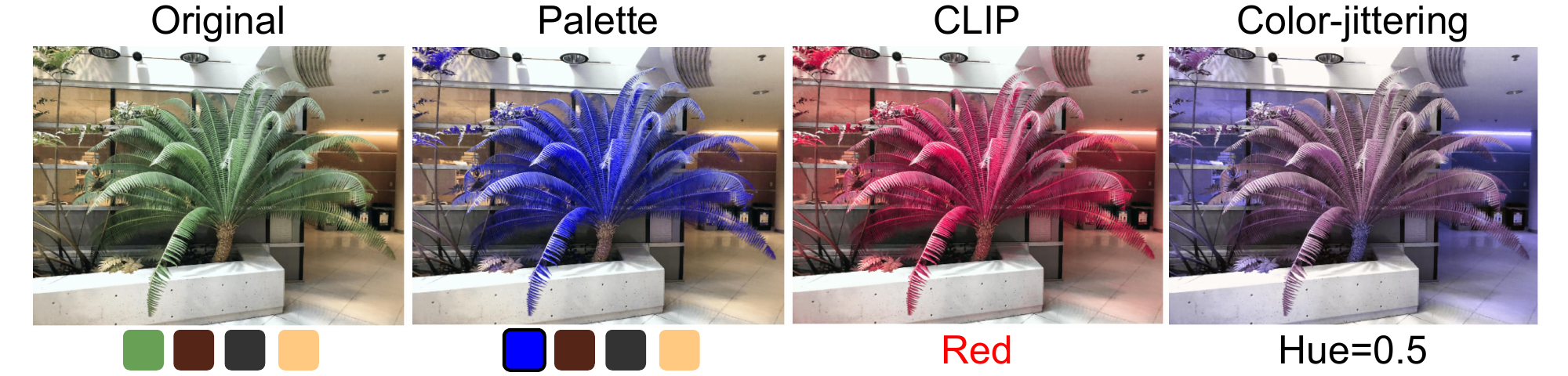}
    \label{fig:model-level-recoloring}
  \end{subfigure}
  
  \caption{Recolorized samples from our selected approaches. From left to right: original image, the result obtained via pallette-based recolorization \cite{Kuang_Luan_Bi_Shu_Wetzstein_Sunkavalli_2022} by changing the green base color to blue, the CLIP-based recolorization \cite{Wang_Chai_He_Chen_Liao_2022} result by giving ``red'' text prompt, and the color-jittering result by changing hue.}
  \label{fig:nerf_recoloring}
\end{figure}

\section{Experiments}
\label{sec:experiments}

\subsection{Experimental settings}
\label{ssec:experimental_settings}

\noindent\textbf{Dataset}. 
We use  two benchmark datasets \textbf{Blender}~\cite{mildenhall2021nerf} and \textbf{LLFF}~\cite{mildenhall2019local} for evaluation. 
For Blender, we directly follow the dataset splitting to use $100$ viewpoints from the training set to train our GeometrySticker and then render $200$ views from the testing set to validate whether the binary messages can be extracted under different viewpoints and color editing conditions. For LLFF, we follow the dataset splitting in NeRF\cite{mildenhall2021nerf}. In general, $1/8$ images in each scene are used to test the visual quality and bit accuracy of binary message extraction, and others are used to train our GeometrySticker.
We report average values across all testing viewpoints in our experiments.
All testing viewpoints are used for computing the average values during the evaluation session.

\noindent\textbf{Baselines.} 
Our evaluations consist of three parts to verify \textbf{claimability}, \textbf{recolorization}, and \textbf{scalability}. For \textbf{claimability}, we compare our proposed GeometrySticker with four baselines for a fair comparison: 1) \textbf{HiDDeN} \cite{zhu2018hidden} $+$ \textbf{NeRF} \cite{mildenhall2021nerf}. We process images with HiDDeN \cite{zhu2018hidden}, a classical image watermarking method, before the training of NeRF; 2) \textbf{CopyRNeRF} \cite{luo2023copyrnerf}. A state-of-the-art method for protecting the copyright of NeRF \cite{mildenhall2021nerf} by using digital watermarking; 3) \textbf{StegaNeRF} \cite{li2023steganerf}. A state-of-the-art data hiding method for steganographic information embedding of NeRF. We adapt StegaNeRF \cite{li2023steganerf} to embed binary messages with the CNN-based message extractor for message retrieval; 4) \textbf{Unwatermarked NeRF}. We also compare the rendered results from NeRF watermarked by GeometrySticker with the rendered results from the non-watermarked version of NeRF to evaluate whether GeometrySticker undermines visual quality. 
For \textbf{recolorization}, we compare the differences between the watermarked recolorized images with the corresponding unwatermarked recolorized images to investigate whether our GeometrySticker undermines recolorization.
For \textbf{scalability}, we validate whether GeometrySticker can be easily adapted into various NeRF architectures, including vanilla NeRF \cite{mildenhall2021nerf}, InstantNGP \cite{mueller2022instant}, and TensoRF \cite{Chen2022ECCV}. We also evaluate whether GeometrySticker is compatible with the existing recolorization schemes mentioned in~\Sref{ssec:recolorization}. 

\noindent\textbf{Evaluation methodology.} 
We evaluate the performance of GeometrySticker compared with other digital watermarking methods using the standard of capacity, invisibility, and robustness. For \textit{capacity}, we set hidden messages bit length to $48$ bits, aligning with the maximum length previously employed in 3D model watermarking methods~\cite{yoo2022deep, luo2023copyrnerf}. For \textit{invisibility}, we evaluate the visual quality with PSNR, SSIM, and LPIPS \cite{Zhang_Isola_Efros_Shechtman_Wang_2018} by comparing the visual quality of the rendered images before and after GeometrySticker watermarking. For \textit{robustness}, we evaluate whether the hidden messages can keep consistent against various distortions and recolorizations. Besides normal situations, we consider different distortions including Gaussian noise, rotation, cropout, and Gaussian blur. Different recolorization pipelines are employed to ensure adequate comparisons.

\begin{table*}[tb]
\centering
\caption{
Reconstruction qualities and message decoding bit accuracies under various recolorizations.
PSNR/SSIM and LPIPS are computed between the watermarked recolorized images and the corresponding unwatermarked recolorized images.
$\uparrow$($\downarrow$) means higher (lower) is better.
``N.A.'' means Not Applicable since CopyRNeRF \cite{luo2023copyrnerf} and StegaNeRF \cite{li2023steganerf} are not compatible with palette-based recolorization. 
}
\label{table:recolorization-robustness}
\begin{adjustbox}{width=0.90\textwidth}
\begin{tabular}{c|c|cc|ccc}
\toprule
\multirow{2}{*}{Datasets} & \multirow{2}{*}{Methods} & \multirow{2}{*}{PSNR/SSIM$\uparrow$} & \multirow{2}{*}{LPIPS$\downarrow$} & \multicolumn{3}{c}{Bit accuracy  $\uparrow$ (\%)} \\
               &  &  & & Color-jitter & CLIP & Palette \\ \midrule
\multirow{4}{*}{Blender} & HiDDeN \cite{zhu2018hidden} $+$ NeRF \cite{mildenhall2021nerf} & 30.80/0.9999 & 0.0167 & 50.13 & 51.08 & 50.91 \\ 
                      & CopyRNeRF \cite{luo2023copyrnerf} & 29.99/0.9999 & 0.0171 & 51.32 & 49.96 & N.A. \\ 
                      & StegaNeRF \cite{li2023steganerf}  & 31.48/0.9999 & 0.0149 & 54.18 & 52.48 & N.A. \\ 
                      & \textbf{GeometrySticker}  & $\mathbf{32.13/0.9999}$ & \textbf{0.0136} & \textbf{99.33} & \textbf{99.50} & \textbf{99.40} \\    
\midrule
\multirow{4}{*}{LLFF} &  HiDDeN \cite{zhu2018hidden} $+$ NeRF \cite{mildenhall2021nerf} & 28.55/0.9999 & 0.2329 & 50.74 & 50.28 & 50.56 \\ 
                         & CopyRNeRF \cite{luo2023copyrnerf} & 27.40/0.9998 & 0.2322 & 47.64 & 49.89 & N.A. \\ 
                         & StegaNeRF \cite{li2023steganerf}  & 29.15/0.9998 & 0.2242 & 52.51 & 50.51 & N.A.\\ 
                         & \textbf{GeometrySticker} & \textbf{30.82/0.9999} & \textbf{0.2165} & \textbf{96.50} & \textbf{97.12} & \textbf{96.88} \\    
\bottomrule
\end{tabular}

\end{adjustbox}
\end{table*}


  

\subsection{Experimental results}
\label{sec:model-color-robustness}

\noindent \textbf{Can we claim ownership over recolorized NeRF models?} 
We first assess whether our GeometrySticker can maintain its effectiveness under various recolorization operations. We consider model-level recolorization, including CLIP-based and palette-based recolorization.  We also apply color jittering to randomly change images' hues as a general image-level color alternation.  As shown in~\Tref{table:recolorization-robustness},  HiDDeN \cite{zhu2018hidden} $+$ NeRF \cite{mildenhall2021nerf} completely fails to perform well under both image- and model-level recolorization. CopyRNeRF \cite{luo2023copyrnerf} also shows degraded performance since CopyRNeRF \cite{luo2023copyrnerf} uses the color representation for hiding the binary message. Since StegaNeRF \cite{li2023steganerf} depends on the complete geometry and color representation for data hiding, the hidden messages are susceptible to being compromised under both image- and model-level recolorization. 
Moreover, the intrinsic architectures of CopyRNeRF \cite{luo2023copyrnerf} and StagaNeRF \cite{li2023steganerf} do not fully support palette-based recolorization, which leads to ``N.A.'' in \Tref{table:recolorization-robustness}. 
In contrast, our approach is uniquely adaptable to all three existing recolorization schemes, further underscoring the scalability of our method. In contrast, the binary messages embedded by our GeometrySticker remain effective against both image- and model-level recolorization and achieve high bit accuracy.


\begin{table*}[tb]
\centering
\caption{
Reconstruction qualities and bit accuracy compared with different baselines.
PSNR/SSIM and LPIPS are computed between the original and watermarked rendered images.
The results are computed on the average of all samples.
}
\label{table:distortion-attacks}
\begin{adjustbox}{width=1.0\textwidth}
\begin{tabular}{c|c|cc|ccccccccc}
\toprule
\multirow{3}{*}{Datasets} & \multirow{3}{*}{Methods} & \multirow{3}{*}{PSNR/SSIM$\uparrow$} & \multirow{3}{*}{LPIPS$\downarrow$} & \multicolumn{5}{c}{Bit accuracy $\uparrow$ (\%)} \\
 &  &  &  & None & Noise & Rotation & Cropout & Blur \\
       &               &      &      &       &  $(\nu=0.1)$   & $(\alpha =\pm \pi / 6)$ & $(s\leq 25 \%)$ & $(\xi=0.1)$ \\
\midrule
\multirow{4}{*}{Blender} & HiDDeN \cite{zhu2018hidden} $+$ NeRF \cite{mildenhall2021nerf} & 30.44/0.9490 & 0.0829 &  50.19  & 49.84 & 50.12 & 50.09 & 50.16 \\ 
                         & CopyRNeRF \cite{luo2023copyrnerf}  & 30.29/0.9478 & 0.0813 &  66.80  & 65.92 & 64.52 & 63.44 & 66.22 \\ 
                         & StegaNeRF \cite{li2023steganerf}  & 30.96/0.9583 & 0.0564  & 100 & 90.21 & 57.17 & 60.30 & 92.88 \\ 
                         & \textbf{GeometrySticker} & \textbf{32.39}/\textbf{0.9713} & \textbf{0.0503} &  \textbf{100}    & \textbf{99.25} & \textbf{98.87} & \textbf{98.75} & \textbf{99.88} \\ 
\midrule
\multirow{4}{*}{LLFF}    & HiDDeN \cite{zhu2018hidden} $+$ NeRF \cite{mildenhall2021nerf} & 23.80/0.7670 & 0.2515 &  51.18  & 49.20 & 50.59 & 50.33 & 49.98 \\ 
                         & CopyRNeRF \cite{luo2023copyrnerf}  & 24.03/0.7747 & 0.2575 &  66.07  & 65.23 & 64.83 & 63.06 & 65.35 \\ 
                         & StegaNeRF \cite{li2023steganerf}  & 24.96/0.8011 & 0.2498 & \textbf{100} & 88.36 & 63.32 & 61.03 & 90.48 \\ 
                         & \textbf{GeometrySticker} & \textbf{25.67}/\textbf{0.8023} & \textbf{0.2465} & \textbf{100} & \textbf{99.36} & \textbf{98.55} & \textbf{98.94} & \textbf{99.59} \\ 
\bottomrule
\end{tabular}
\end{adjustbox}
\end{table*}

\begin{figure*}[t!]  
  \centering
  \includegraphics[width=1.0\linewidth]{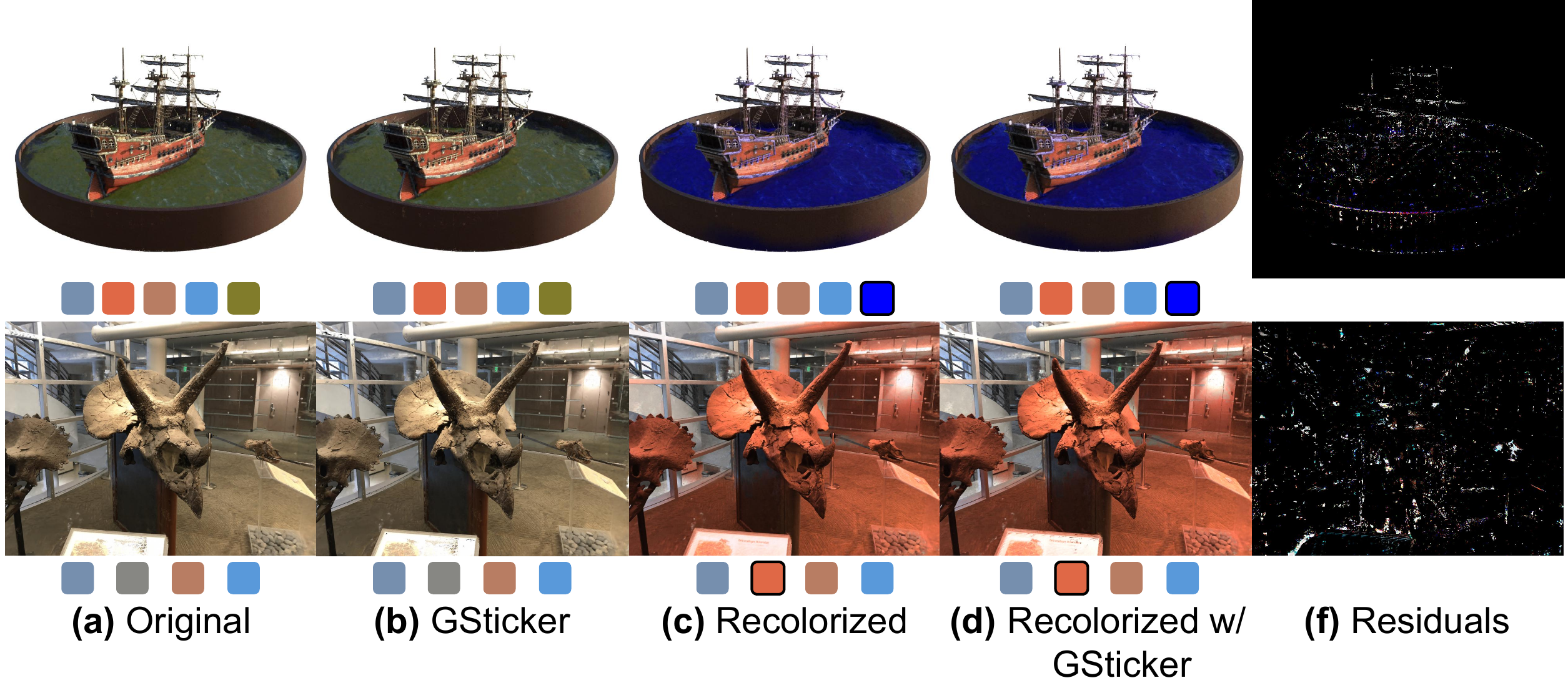}


  \caption{Qualitative evaluation of whether GeometrySticker (GSticker) compromises the recolorization. 
  Column \textbf{(a)} is rendered from an original NeRF without GSticker. 
  Column \textbf{(b)} is rendered from NeRF watermarked by GeometrySticker. 
  Column \textbf{(c)} is the result of applying palette-based recolorization on NeRF models in Column \textbf{(a)}.
  Column \textbf{(d)} is the result of applying palette-based recolorization on NeRF models in Column \textbf{(b)}.
  Column \textbf{(e)} shows \textbf{residual maps} between Column \textbf{(c)} and Column \textbf{(d)}. 
  The minor differences in the residual map indicate that GeometrySticker does not undermine the recolorization.
  }
  \label{fig:residual-maps}
\end{figure*}

\noindent \textbf{Does GeometrySticker undermine recolorization?} We discuss whether our GeometrySticker undermines recolorization in \Tref{table:recolorization-robustness} and \Fref{fig:residual-maps}. We evaluate the variations between recolored samples from NeRF models that are not watermarked and those that have been watermarked. In \Tref{table:recolorization-robustness}, as it is difficult to obtain identical recolorized pairs via CLIP-based recolorization, we use Palette-based colorization for our approach and HiDDeN \cite{zhu2018hidden} $+$ NeRF\cite{mildenhall2021nerf}. We utilize color-jittering to change the images' hue for CopyRNeRF \cite{luo2023copyrnerf} and StegaNeRF \cite{li2023steganerf}, since the two methods are not compatible with Palette-based colorization. The better quantitative values in \Tref{table:recolorization-robustness} show that our samples can achieve higher similarity to the unwatermarked recolorized images. This is further supported by the qualitative results shown in \Fref{fig:residual-maps}. Moreover, our technique uniquely maintains a balance between recolorization quality and bit accuracy post-recolorization, unlike other methods, which show significantly reduced bit accuracy following recolorization.

\begin{table}[tb]
   \centering
   \caption{Evaluation of scalability by incorporating our GeometrySticker into different NeRF architectures. PSNR/SSIM and LPIPS are computed between the samples rendered from watermarked NeRF models and the corresponding ground truth images without watermarks. 
   }

    \begin{adjustbox}{width=0.6\textwidth}
   \begin{tabular}{c|ccc}
    \toprule
    Architecture  & PSNR/SSIM$\uparrow$ & LPIPS$\downarrow$ & Bit accuracy $\uparrow$ \\
    \midrule
     NeRF\cite{mildenhall2021nerf} & 27.44/0.8759 & 0.1667 & 100\%  \\ 
    InstantNGP\cite{mueller2022instant}  & 28.59/0.8868 &  0.1304 & 100\% \\
    TensoRF\cite{Chen2022ECCV}     &  29.18/0.8907  & 0.1370 &  100\% \\
    \bottomrule
  \end{tabular}
  \end{adjustbox}
  
  \label{tab:scalability}
\end{table}

\noindent \textbf{Can GeometrySticker function properly without recolorization?} We evaluate if GeometrySticker functions properly in standard scenarios without recolorization. We also evaluate its robustness by applying several types of 2D distortions to rendered images, including Gaussian noise with deviation $\nu$, random rotation with parameters $\alpha$, random cropout with a parameter $s$, and Gaussian blur with deviation $\xi$.  As shown in the \Tref{table:distortion-attacks}, HiDDeN \cite{zhu2018hidden} fails to extract the binary messages from the renderings of NeRF. 
CopyRNeRF \cite{luo2023copyrnerf} can limitedly extract hidden messages from the renderings and show undermined robustness to different image distortions.
Although StegaNeRF \cite{li2023steganerf} can extract the hidden messages, it shows vulnerability to different types of image distortions. Our GeometryStricker reliably extracts hidden messages and shows robustness to different image distortions.

\noindent \textbf{Is GeometrySticker scalable?} We have shown that our method can achieve scalability over the three main recolorization schemes in~\Tref{table:recolorization-robustness}. We further evaluate the scalability of our proposed GeometrySticker on three typical NeRF architectures, including vanilla NeRF \cite{mildenhall2021nerf}, InstantNGP \cite{mueller2022instant}, and TensoRF \cite{Chen2022ECCV}. From~ \Tref{tab:scalability}, GeometrySticker can achieve high invisibility and bit accuracy in these conditions, which reconfirms our scalability.

\begin{figure*}[t!]
  \centering
   \includegraphics[width=\linewidth]{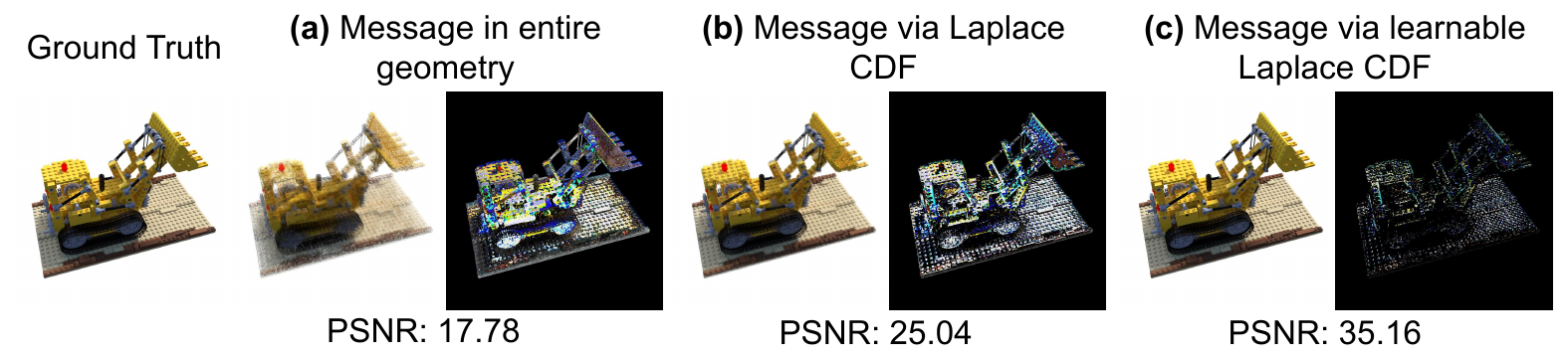}

  \caption{Ablation study to our cover media selection. \textbf{(a)} indicates including all geometry points for messages sticking, which can cause obvious distortion on the scene surface.  \textbf{(b)} indicates using Laplace CDF with a fixed threshold for message attachment, which reduces perturbation on the NeRF geometry but still shows noticeable distortion. \textbf{(c)} indicates using our learnable Laplace CDF can find an optimal threshold, making the visual distortion almost imperceivable. The visual quality results of PSNR are averaged on all examples in the selected scene.}
  \label{fig:ablation_study}
\end{figure*}

\noindent\textbf{Ablation study}. The selection strategy of 3D points is a key architecture in our framework. We focus on investigating this part in our ablation study. As shown in \Fref{fig:ablation_study}, attaching messages to all geometry components can cause obvious distortion, which is aligned with the previous findings in CopyRNeRF \cite{luo2023copyrnerf}. Applying simple Laplace CDF with fixed thresholds for message attachment can reduce perturbation on the NeRF geometry but still with noticeable distortion. Our learnable Laplace CDF can find an optimal threshold for message attachment, making the visual distortion imperceivable.

\noindent\textbf{Potential threat analysis}. The experiments in~\Tref{table:distortion-attacks} have highlighted the robustness of our method to common image distortions. Besides, as the recolorization is also a very powerful modification operation, our previous experiments also demonstrate that GeometrySticker can show robustness to different recolorizations. We further investigate the robustness of the embedded watermarks against various possible deliberate interferences and security threats including the adversarial attack and model purification.


\captionsetup[figure]{font=small,skip=0pt}
\begin{wrapfigure}{r}{0.48\textwidth}
  \vspace{-8mm}
  \begin{center}
    \includegraphics[width=1.0\textwidth]{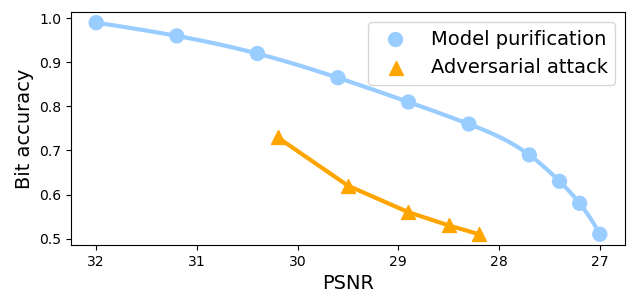}
  \end{center}
  \caption{ Robustness to model purification and adversarial attacks. We show the correlations between the PSNR and bit accuracy.}
  \label{fig:threat_model_analysis}
  \vspace{-7mm}
\end{wrapfigure}

\noindent \textit{Adversarial attack.}
We consider the situation if the message extractor has leaked. A malicious user can use an adversarial attack such as PGD \cite{madry2017towards} to remove the hidden messages by optimizing the rendered images via a PSNR constraint. 
The goal is to minimize the Euclidean distance between a pre-sampled random binary message and the message extractor's output, which could replace the original hidden message with a random one. As shown in \Fref{fig:threat_model_analysis}, adversarial attacks can indeed result in a reduction of bit accuracy while causing only minimal visual distortion. 
Thus, it's essential to keep the message extractor private.

\noindent \textit{Model purification.}
Another one is the model purification by fine-tuning the model with non-watermarked images. We consider an extreme situation that the attackers can directly access the original non-watermarked images used for the NeRF creation.  
Based on this assumption, we implement this attack on GeometrySticker by eliminating the message loss and fine-tuning the model solely through perceptual loss. 

As shown in \Fref{fig:threat_model_analysis}, the bit accuracy starts to decrease if the model purification sacrifices the rendering quality, but it can keep a relatively high bit accuracy if we want to keep the rendered image quality.
These results show that model purification is hard to reduce the bit accuracy significantly without sacrificing the image quality.



\section{Conclusion}
In our research, we introduce a novel approach for asserting ownership of recolorized NeRF models through the innovative GeometrySticker. By embedding binary messages within the high geometry value components of NeRF, we ensure that these messages remain robust even after recolorization. 
Comprehensive testing demonstrates that our method establishes ownership claims on recolorized NeRF models, which guarantees the safe application of NeRF recolorization across various scenarios, thereby ensuring positive societal impacts for protecting the copyrights of artists and creators.

\noindent\textbf{Limitations and future work}. Our method is an effective technical solution for copyright protection against the recolorization of NeRF models. However, as we discussed before, our mechanism may still face threats from some malicious operations. 
We will consider further enhancing the adversary robustness through adversary learning approaches \cite{madry2017towards, pang2024heterogeneous, li2024graph}.
Besides, we will explore enhancing the robustness of GeometrySticker towards geometry editing scenarios, 
enabling the manipulation of NeRF model and rendered images through techniques such as cage-based deformation \cite{xu2022deforming} or motion transfer \cite{huang2024motion} in future work.
Moreover, we will improve the GeometrySticker to align with the emerging 3D Gaussian Splatting (3DGS) method \cite{kerbl20233d}, which utilizes an explicit point cloud representation, distinguishing itself from the NeRF implicit neural representation.
Our aim is to elevate the versatility of GeometrySticker to be compatible with different 3D representation baselines.

\noindent\textbf{Acknowledgement}
This work was done at Renjie’s Research Group at the Department of Computer Science of Hong Kong Baptist University. Renjie's Research Group is supported by the National Natural Science Foundation of China under Grant No. 62302415, Guangdong Basic and Applied Basic Research Foundation under Grant No. 2022A1515110692, 2024A1515012822, and the Blue Sky Research Fund of HKBU under Grant No. BSRF/21-22/16.


%
%
\bibliographystyle{splncs04}
\bibliography{main}

\title{Supplementary Material: \\ GeometrySticker: Enabling Ownership Claim of Recolorized Neural Radiance Fields} 

\titlerunning{Enabling Ownership Claim of Recolorized Neural Radiance Fields}

\author{Xiufeng Huang\inst{1,2} \and 
Ka Chun Cheung\inst{2} \and 
Simon See\inst{2} \and 
Renjie Wan\inst{1}\thanks{Corresponding author.}
}

\authorrunning{X.~Huang et al.}

\institute{Department of Computer Science, Hong Kong Baptist University \and
NVIDIA AI Technology Center, NVIDIA \\
\email{xiufenghuang@life.hkbu.edu.hk, \{chcheung, ssee\}@nvidia.com, renjiewan@hkbu.edu.hk}}

\maketitle

\section{Overview}
\label{sec:overview}
This supplementary document provides more discussions, implementation details, and further results that accompany the paper:

\begin{itemize}

    \item \Sref{sec:uniqueness} explains the uniqueness of our method by comparing with the current NeRF ownership claiming methods under NeRF recolorizations.

    \item \Sref{sec:laplace_cdf} explains the effectiveness of applying the Laplace Cumulative Distribution Function (CDF) with learnable parameters.
    

    \item \Sref{sec:recolorization_methods} introduces the details of our reference colors and visualizes their corresponding recolorization results for NeRF. These recolorization methods are applied to different NeRF architectures to validate ownership for the recolorized NeRF.
    
    \item \Sref{sec:implementation_details} presents the implementation details of our method, including the network architectures and the training process.

    \item \Sref{sec:additional_results} provides additional results, including additional qualitative results of the main paper.
    
\end{itemize}

\section{Uniquesness}
\label{sec:uniqueness}
\begin{figure}[t!]
  \centering
  \begin{subfigure}{\linewidth}
    \includegraphics[width=\linewidth]{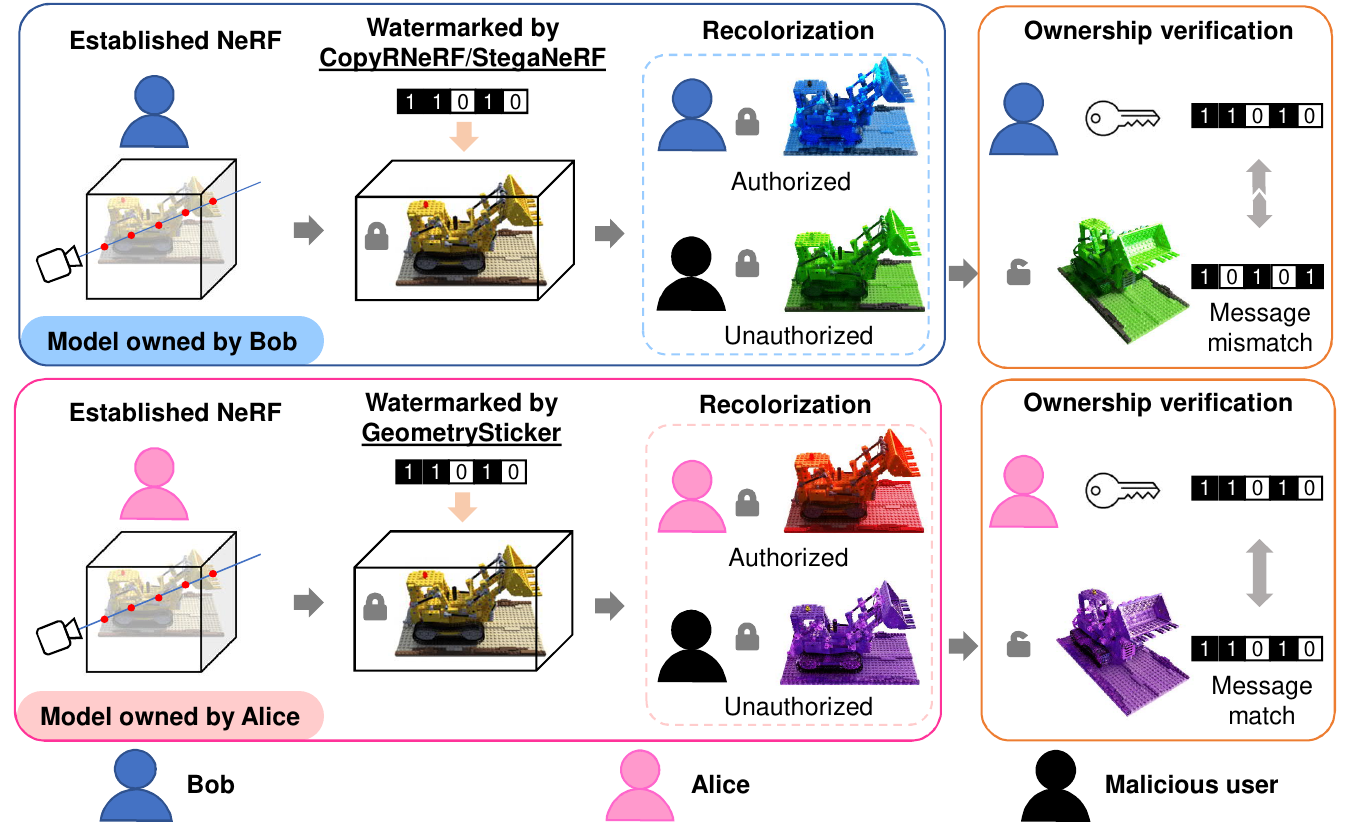}
    \label{fig:model-level-recoloring}
  \end{subfigure}
  
  \caption{Illustration of the uniqueness of our method. 
  The first row illustrates the NeRF model owner \textit{Bob} claims the ownership by using CopyRNeRF \cite{luo2023copyrnerf} or StegaNeRF \cite{li2023steganerf}. However, when a malicious user applies unauthorized recolorization on \textit{Bob}'s model, the hidden ownership information can be corrupted and mismatch the original secret messages.
  The second row illustrates the NeRF model owner \textit{Alice} claims the ownership by GeometrySticker. The NeRF model watermarked by GeometryStricker can be robust to different recolorizations. Even if a malicious user applies unauthorized recolorization on \textit{Alice}'s model, the hidden ownership information can still be reliably extracted and match the original secret messages.
  }
  \label{fig:uniqueness}
\end{figure}

As shown in \Fref{fig:uniqueness}, we demonstrate the uniqueness of using our GeometrySticker to claim ownership of a recolorized NeRF model. 
The current ownership protection methods such as CopyRNeRF \cite{luo2023copyrnerf} and StegaNeRF \cite{li2023steganerf} can only claim the ownership when recolorization is not conducted.
However, since the recent developments of NeRF recolorization methods \cite{Wang_Chai_He_Chen_Liao_2022, Kuang_Luan_Bi_Shu_Wetzstein_Sunkavalli_2022, gong2023recolornerf}, if a model owner \textit{Bob} creates a NeRF model and watermark the model with CopyRNeRF \cite{luo2023copyrnerf} or StegaNeRF \cite{li2023steganerf}, the hidden ownership information could be vulnerable when a malicious user applies unauthorized recolorization on the NeRF model. 
Our GeometrySticker can be robust under different recolorizations.
A model owner \textit{Alice} can watermark her NeRF model by GeometrySticker, which can keep the hidden information consistent under different recolorizations and reliably extract the binary message from the recolorized NeRF renderings. 

\section{Learnable Laplace CDF}
\label{sec:laplace_cdf}

\begin{figure}[t!]
  \centering
  \begin{subfigure}{\linewidth}
    \includegraphics[width=\linewidth]{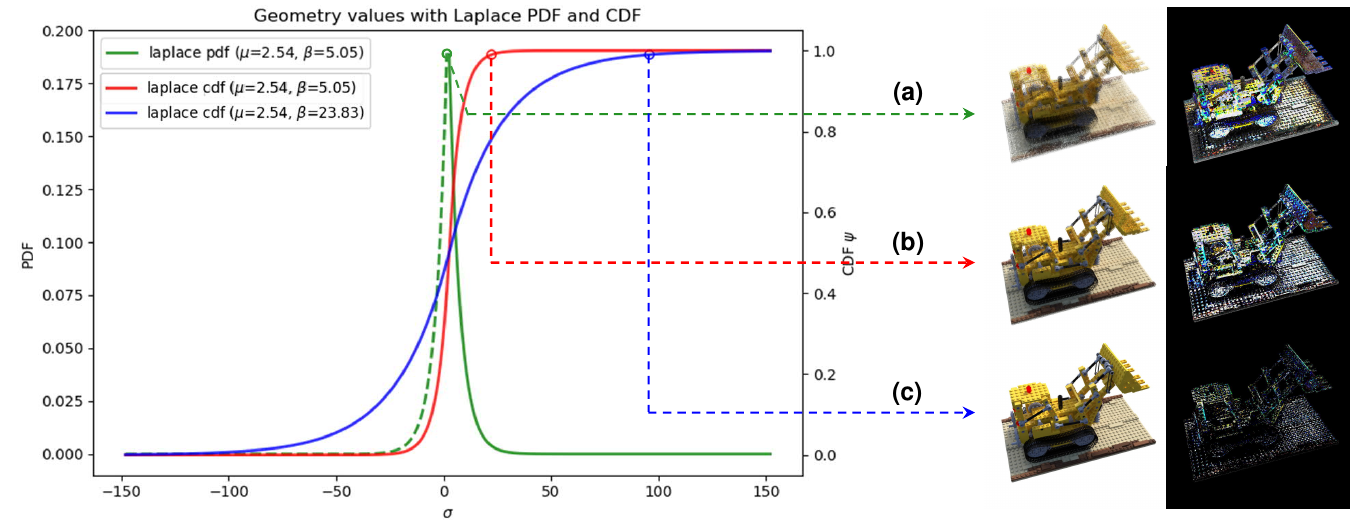}
    \label{fig:model-level-recoloring}
  \end{subfigure}
  
  \caption{Message attachment into NeRF geometry values by applying Laplace CDF with different deviation parameters. 
  The geometry values distribution is modeled by a Laplace distribution with the mean $\mu$ and deviation $\beta$. 
  \textbf{(a)} indicates directly attaching messages on all geometry values can cause obvious distortion. \textbf{(b)} indicates applying Laplace CDF with fixed $\mu$ and $\beta$ can reduce perturbation but still show noticeable distortion. \textbf{(c)} indicates applying Laplace CDF with a learnable deviation parameter $\beta$ can find an optimized threshold for filtering 3D points and make the distortion imperceivable.}
  \label{fig:laplace_cdf}
\end{figure}

We provide more ablation studies for our learnable Laplace CDF used for the selection of cover medium.
As shown in \Fref{fig:laplace_cdf}, 
we calculate the mean $\mu$ and deviation $\beta$ of the geometry values and use the Laplace distribution to model the geometry values distribution of a selected scene.
As shown in \Fref{fig:laplace_cdf} \textbf{(a)}, attaching messages to all NeRF geometry values can cause obvious distortion since the low geometry values take up the majority of the entire NeRF geometry. 
We apply the Laplace CDF with the fixed parameters $\mu$ and $\beta$ and the CDF value $\psi=0.99$ as the threshold to filter large geometry values for messages attachment.
As shown in \Fref{fig:laplace_cdf} \textbf{(b)}, applying Laplace CDF with calculated parameters can reduce perturbation but still show noticeable distortion. 
As shown in \Fref{fig:laplace_cdf} \textbf{(c)}, our learnable Laplace CDF can adaptatively find an optimized deviation parameter $\beta$ to adjust the CDF threshold ($\psi=0.99$) for the selection of cover medium and finally make the perturbation caused by the attached messages imperceivable.

\section{More details on recolorization}
\label{sec:recolorization_methods}

\begin{figure}[p]
  \centering

  \begin{subfigure}{\linewidth}
    \includegraphics[width=\linewidth]{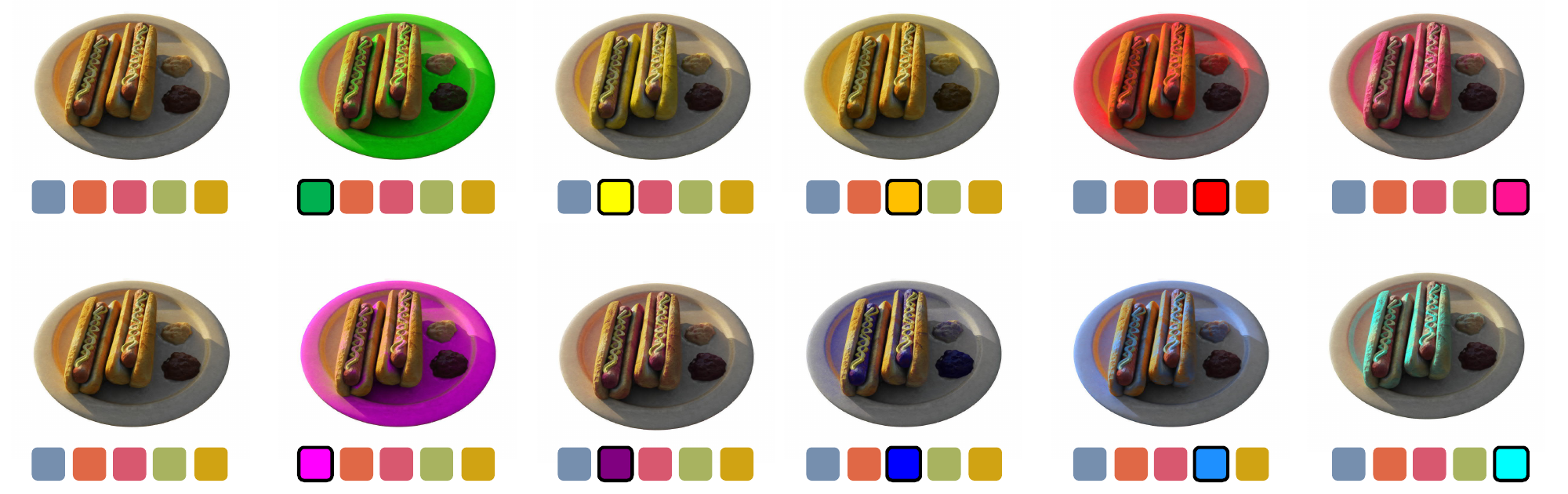}
    \vspace{-3mm}
    \caption{Palette-based recolorization}
    \label{fig:palette-based-recolorization}
  \end{subfigure}

  \begin{subfigure}{\linewidth}
    \includegraphics[width=\linewidth]{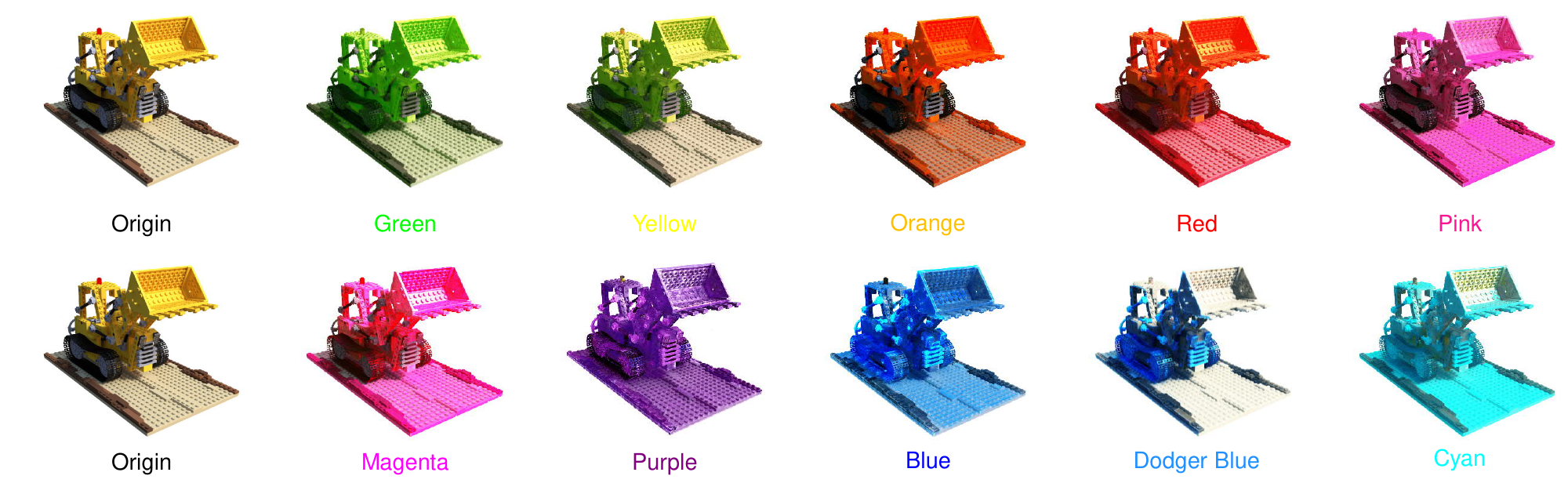}
    \vspace{-3mm}
    \caption{CLIP-based recolorization}
    \label{fig:clip-based-recolorization}
  \end{subfigure}

  \begin{subfigure}{\linewidth}
    \includegraphics[width=\linewidth]{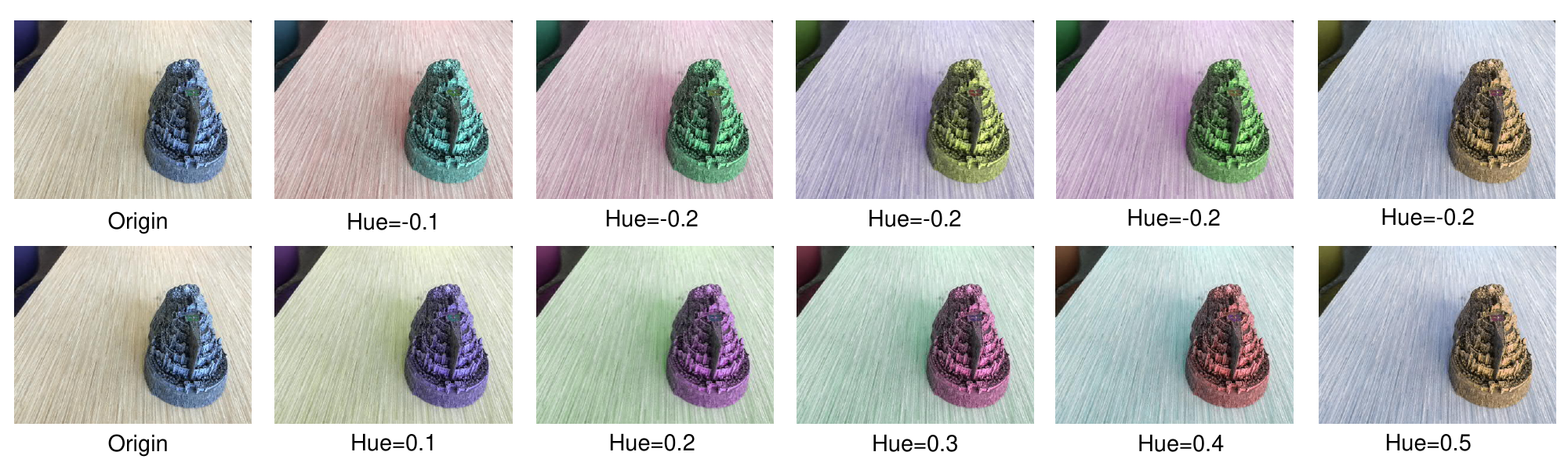}
    \vspace{-3mm}
    \caption{Color-jittering recolorization}
    \label{fig:color-jittering}
  \end{subfigure}

  \begin{subfigure}{\linewidth}
    \includegraphics[width=\linewidth]{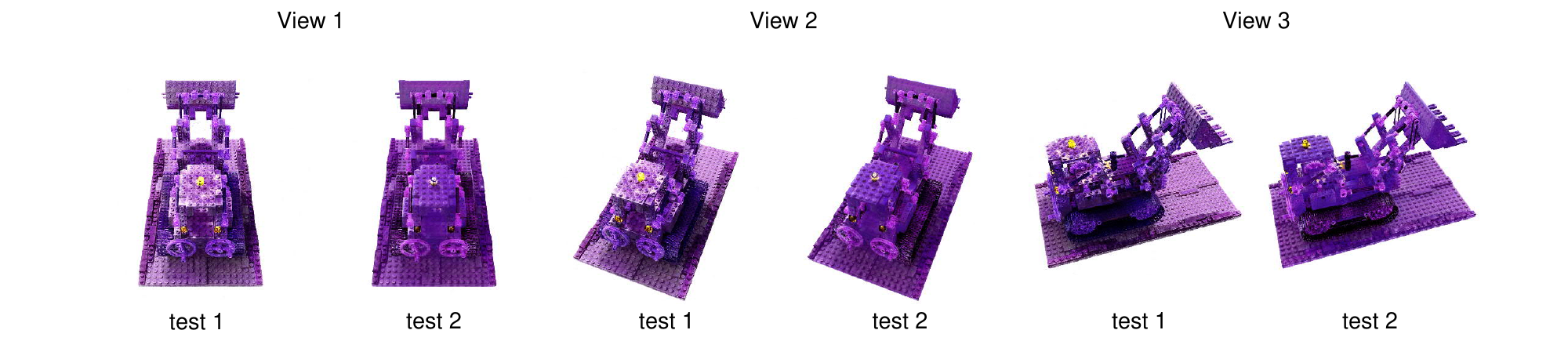}
    \vspace{-3mm}
    \caption{CLIP-based recolorization with the same text prompt "purple" can have different results.}
    \label{fig:color-clip-diff}
  \end{subfigure}
  
  \caption{Recolorization results by using different methods: \textbf{(a)} is the Palette-based recolorization. The first column is the reference image with the original color palette, and others are recolorized by assigning reference colors to different base colors in the color palette. \textbf{(b)}  is the CLIP-based recolorization. The first column is the reference image, and the others are recolorized by using the colors' name as the text prompt. \textbf{(c)} is the color-jittering recolorization. The first column is the original image and others are recolorized by changing the hue of the original image by shifting the intensities with the range of $[-0.5, 0.5]$ in the hue channel. \textbf{(d)} indicates CLIP-based recolorization with the same text prompt can have different results.}
  \label{fig:recolorizations}
\end{figure}

We select $10$ reference colors from the Standard sRGB / Rec.$709$ color gamut including green, yellow, orange, red, pink, megenta, purple, blue, dodger blue, cyan. We recolorize the NeRF models by using colors' name for the CLIP-based method or assigning RGB values for the palette-based method. We also convert the NeRF renderings into HSV format to recolorize the images by changing the hue channel. As shown in \Fref{fig:palette-based-recolorization}, the palette-based method can precisely recolorize NeRF by editing the palette's colors to the reference colors. As shown in \Fref{fig:clip-based-recolorization}, 
though the CLIP-based method can roughly conduct the recolorization via the text prompts, the results are uncontrollable since the recolorization under the same prompts may have some differences as shown in \Fref{fig:color-clip-diff}.
Thus, it is hard to get the same results for an unwatermarked NeRF model and a watermarked NeRF model. As shown in \Fref{fig:color-jittering}, color-jittering is an image-level recolorization by converting images into HSV format and shifting the intensity of the hue channels in a scale of $[-0.5, 0.5]$.
For a fair comparison across different baselines, we only use color-jittering in our reconstruction quality computation for PSNR/SSIM and LPIPS in the main manuscript Table \textcolor{red}{1}, since CLIP-based recolorization is uncontrollable and palette-based recolorization is not applicable to CopyRNeRF \cite{luo2023copyrnerf} and StegaNeRF \cite{li2023steganerf}. All the testing set images in the main manuscript Section \textcolor{red}{5.1} are recolorized for computing reconstruction quality or message extraction bit accuracies.

\section{Implementation details}
\label{sec:implementation_details}

\subsection{Network achitectures}
In our proposed GeometrySticker, the message sticker $\Theta_{\mathbf{m}}$ is an MLP layer. In specific, it has $80$ input channels, which are a concatenation of the message $\mathbf{M}$ in $48$ dimensions and positional encoding $\gamma_{x}(\mathbf{x})$ in $32$ dimensions. The message sticker $\Theta_{\mathbf{m}}$ has two hidden layers with 64 dimensions and 1-dimensional output for the message embedding $m$. 
For the message extractor $D_{\chi}$, we use the VGG16 network \cite{simonyan2014very} as the backbone feature extractor. An average pooling is then performed, followed by a final linear layer with a fixed output dimension $N_{b}$ to produce the continuous predicted message $\hat{\mathbf{M}}$.
For the watermark classifier $C_{\phi}$, we use a similar architecture with the message extractor $D_{\chi}$ with the VGG16 network \cite{simonyan2014very} as the feature extractor followed by an average pooling layer and a final 1-dimensional layer for classification.

\begin{figure*}[t!]  
  \centering
  \begin{subfigure}{\linewidth}
    \begin{subfigure}{0.49\linewidth}
      \includegraphics[width=\linewidth]{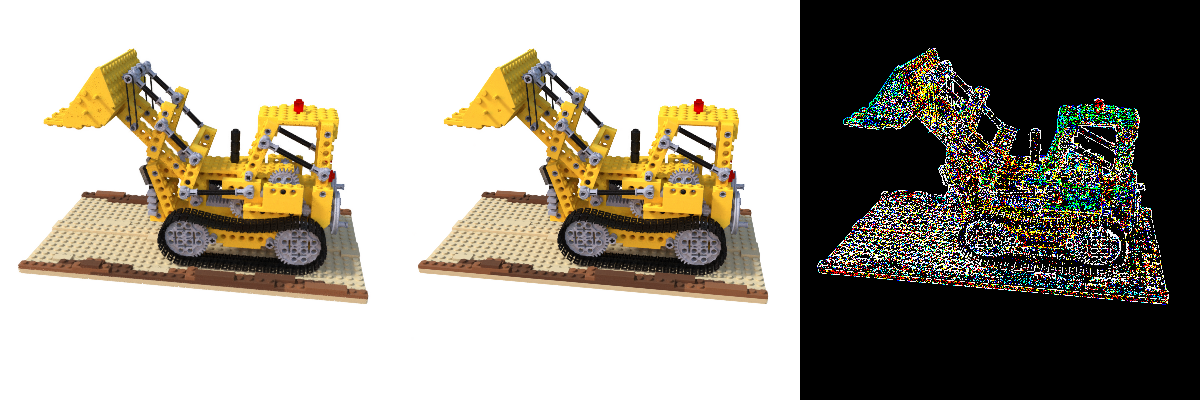}
      \caption{PSNR=34.41, bit accuracy=$100\%$}
    \end{subfigure}
    \begin{subfigure}{0.49\linewidth}
      \includegraphics[width=\linewidth]{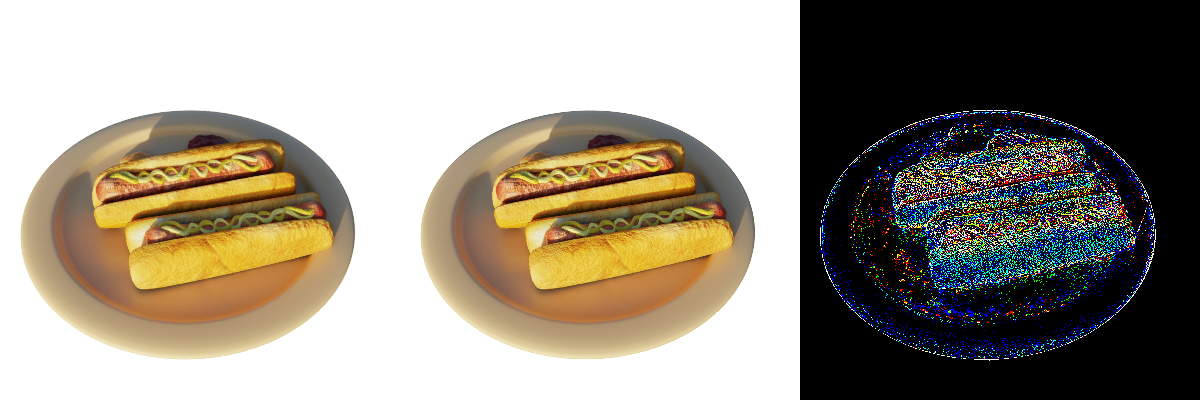}
      \caption{PSNR=36.91, bit accuracy=$100\%$}
    \end{subfigure}
  \end{subfigure}

  \begin{subfigure}{\linewidth}
    \begin{subfigure}{0.49\linewidth}
      \includegraphics[width=\linewidth]{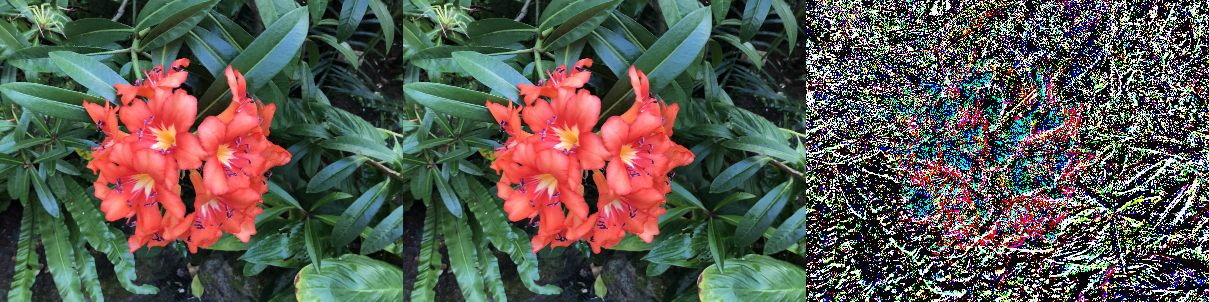}
      \caption{PSNR=31.62, bit accuracy=$100\%$}
    \end{subfigure}
    \begin{subfigure}{0.49\linewidth}
      \includegraphics[width=\linewidth]{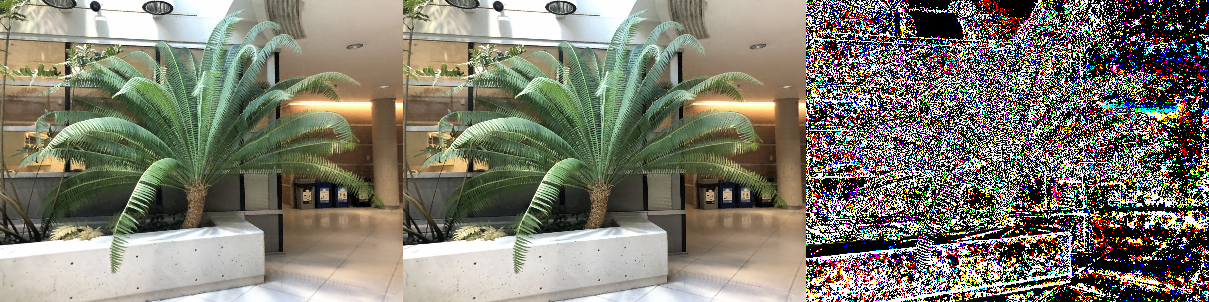}
      \caption{PSNR=30.70, bit accuracy=$100\%$}
    \end{subfigure}
    \label{fig:residual-maps-normal}
  \end{subfigure}

  \caption{Additional results for different scenes. The message length is 48 bits. We visualize the residual maps between the unwatermarked renderings and the watermarked renderings. From left to right: unwatermarked, GeometrySticker, residual maps ($\times 10$).}
  \label{fig:residual-maps-normal}
\end{figure*}

\subsection{Training process}
The training process consists of two stages. In the first stage, we establish a NeRF scene by optimizing $\Theta_{\sigma}$ and $\Theta_{c}$ to get the geometry and color values of the scene according to $\mathcal{L}_{cont}$. 
In the second stage, we keep the geometry MLP $\Theta_{\sigma}$ and color MLP $\Theta_{c}$ unchanged and train the message sticker $\Theta_{m}$ and Laplace CDF with the learnable deviation parameter $\beta$ for message attachment and key points selection.
Meanwhile, we train a message extractor $D_{\chi}$ to extract the hidden message from the 2D watermarked renderings.
In addition, we also train the watermarking classifier $C_{\phi}$ to classify whether the NeRF renderings contain watermarkings or not.
The $\mathcal{L}_{cont}$ is measured by the mean squared error between the watermarked rendering images and the ground truth images.
The $\mathcal{L}_{msg}$ is a binary cross entropy loss calculated between the embedded messages $\mathbf{M}$ and the extracted messages $\mathbf{\hat{M}}$.
The $\mathcal{L}_{cls}$ is a binary cross entropy loss calculated between the watermarked rendering image $\mathbf{I}_{w}$ and the unwatermarked rendering images $\mathbf{I}_{u}$.
$\mathcal{L}_{sparse}$ is the sparsity loss \cite{Lombardi_Simon_Saragih_Schwartz_Lehrmann_Sheikh_2019} to force the CDF value $\psi$ to be close to either zero or one.
The network $\Theta_{m}$ and parameters $\chi$, $\phi$ and $\beta$ are optimized with the objective functions $\mathcal{L}_{cont}$, $\mathcal{L}_{msg}$, $\mathcal{L}_{cls}$ and $\mathcal{L}_{sparse}$.
In every training loop, we attach the message $\mathbf{M}$ with a random camera pose and apply 2D distortions on the watermarked rendering images.

\begin{figure*}[t!]  
  \centering
  \begin{subfigure}{\linewidth}
    \includegraphics[width=\linewidth]{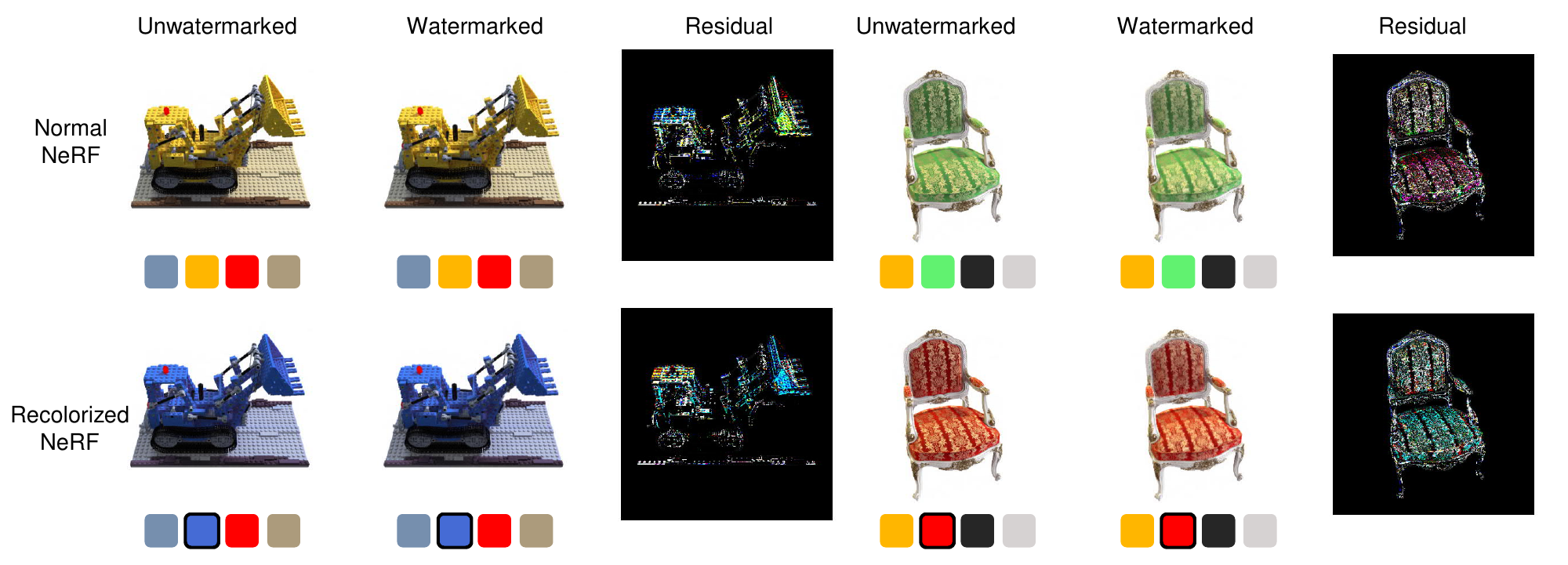}
  \end{subfigure}

  \caption{Residual maps for NeRF renderings before and after palette-based recolorizations. The first row shows the residual maps before palette-based recolorizations. The second row shows the residual maps after palette-based recolorizations. Each residual map shows the differences between the unwatermarked renderings and watermarked renderings by GeometrySticker.}
 \label{fig:residual-maps-recoloring}
\end{figure*}

\section{Additional results}
\label{sec:additional_results}

We provide additional results to validate the effectiveness of our GeometrySticker.
As shown in \Fref{fig:residual-maps-normal}, we evaluate the qualitative and quantitative results of the reconstruction quality and bit accuracies of our GeometrySticker on the selected scene.
The watermarked rendered images have high reconstruction quality with minimal discrepancies compared with the original rendered images. From the residual maps, we can observe that the hidden messages are sparsely embedded into the geometrical structure of the object or scene.

We further validate the consistency of our GeometrySticker under different recolorizations.
As shown in \Fref{fig:residual-maps-recoloring},
the message perturbation attached by GeometrySticker remains consistent from non-recolorized NeRF models to recolorized NeRF models.
These results show our method successfully embeds secret messages into the geometry representation and disentangles them with the color representation, thus claiming ownership under various NeRF recolorizations.


\end{document}